\documentclass[11pt,a4paper]{article}
\usepackage[nohyperref]{emnlp-ijcnlp-2019}
\usepackage{times}
\usepackage{latexsym}

\usepackage{xurl}

\usepackage{amsmath}
\usepackage{booktabs}
\usepackage{bm}
\usepackage{bbm}
\usepackage{multirow}

 \aclfinalcopy 

\DeclareMathOperator*{\argmin}{arg\,min}

\title{Unsupervised Text Summarization via Mixed Model Back-Translation}

\author{Yacine Jernite \\
Facebook AI Research, New York, NY \\
  {\tt yjernite@fb.com} \\}

\date{}

\begin{document}
\maketitle
\begin{abstract}
 Back-translation based approaches have recently lead to significant progress in unsupervised sequence-to-sequence tasks such as machine translation or style transfer. In this work, we extend the paradigm to the problem of learning a sentence summarization system from unaligned data. We present several initial models which rely on the asymmetrical nature of the task to perform the first back-translation step, and demonstrate the value of combining the data created by these diverse initialization methods. Our system outperforms the current state-of-the-art for unsupervised sentence summarization from fully unaligned data by over 2 \textsc{ROUGE}, and matches the performance of recent semi-supervised approaches.
\end{abstract}

\section{Introduction}

Machine summarization systems have made significant progress in recent years, especially in the domain of news text. This has been made possible among other things by the popularization of the neural sequence-to-sequence (seq2seq) paradigm \citep{DBLP:conf/emnlp/KalchbrennerB13,DBLP:conf/nips/SutskeverVL14,DBLP:conf/ssst/ChoMBB14}, the development of methods which combine the strengths of extractive and abstractive approaches to summarization \citep{DBLP:conf/acl/SeeLM17,DBLP:conf/emnlp/GehrmannDR18}, and the availability of large training datasets for the task, such as Gigaword or the CNN-Daily Mail corpus which comprise of over 3.8M shorter and 300K longer articles and aligned summaries respectively.
Unfortunately, the lack of datasets of similar scale for other text genres remains a limiting factor when attempting to take full advantage of these modeling advances using supervised training algorithms.

In this work, we investigate the application of back-translation to training a summarization system in an unsupervised fashion from unaligned full text and summaries corpora. Back-translation has been successfully applied to unsupervised training for other sequence to sequence tasks such as machine translation \citep{DBLP:conf/emnlp/LampleOCDR18} or style transfer \citep{DBLP:journals/corr/SubramanianLSDRB18}.
We outline the main differences between these settings and text summarization, devise initialization strategies which take advantage of the asymmetrical nature of the task, and demonstrate the advantage of combining varied initializers. 
Our approach outperforms the previous state-of-the-art on unsupervised text summarization while using less training data, and even matches the \textsc{rouge} scores of recent semi-supervised methods.

\section{Related Work}

\citet{DBLP:conf/emnlp/RushCW15}'s work on applying neural seq2seq systems to the task of text summarization has been followed by a number of works improving upon the initial model architecture.
These have included changing the base encoder structure \citep{DBLP:conf/naacl/ChopraAR16}, adding a pointer mechanism to directly re-use input words in the summary \citep{DBLP:conf/conll/NallapatiZSGX16,DBLP:conf/acl/SeeLM17}, or explicitly pre-selecting parts of the full text to focus on \citep{DBLP:conf/emnlp/GehrmannDR18}. 
While there have been comparatively few attempts to train these models with less supervision, auto-encoding based approaches have met some success \citep{DBLP:conf/emnlp/MiaoB16,DBLP:conf/emnlp/WangL18a}.

\citet{DBLP:conf/emnlp/MiaoB16}'s work endeavors to use summaries as a discrete latent variable for a text auto-encoder. They train a system on a combination of the classical log-likelihood loss of the supervised setting and a reconstruction objective which requires the full text to be mostly recoverable from the produced summary. While their method is able to take advantage of unlabelled data, it relies on a good initialization of the encoder part of the system which still needs to be learned on a significant number of aligned pairs. \citet{DBLP:conf/emnlp/WangL18a} expand upon this approach by replacing the need for supervised data with adversarial objectives which encourage the summaries to be structured like natural language, allowing them to train a system in a fully unsupervised setting from unaligned corpora of full text and summary sequences. Finally, \citep{DBLP:journals/corr/SongTQLL19} uses a general purpose pre-trained text encoder to learn a summarization system from fewer examples. Their proposed MASS scheme is shown to be more efficient than BERT \citep{DBLP:journals/corr/DevlinCLT18} or Denoising Auto-Encoders (DAE) \citep{DBLP:conf/icml/VincentLBM08,DBLP:conf/aaai/FuTPZY18}.

This work proposes a different approach to unsupervised training based on back-translation. The idea of using an initial weak system to create and iteratively refine artificial training data for a supervised algorithm has been successfully applied to semi-supervised \citep{DBLP:conf/acl/SennrichHB16} and unsupervised machine translation \citep{DBLP:conf/emnlp/LampleOCDR18} as well as style transfer \citep{DBLP:journals/corr/SubramanianLSDRB18}. We investigate how the same general paradigm may be applied to the task of summarizing text.

\section{Mixed Model Back-Translation}

\begin{table*}[t]
    \centering
    \normalsize
    \begin{tabular}{l}
         (Original) \emph{france took an important step toward power market liberalization monday, braving } \\ 
         \emph{union anger to announce the partial privatization of state-owned behemoth electricite de france.} \\
         \bottomrule
        (Pr-Thr)  france launched a partial UNK of state-controlled utility, the privatization agency said. \\
        \bottomrule 
        (DBAE)    france's state-owned gaz de france sa said tuesday it was considering partial partial  \\
                privatization of france's state-owned nuclear power plants. \\
        \bottomrule
        ($\bm{\mu}:1$) france launches an initial public announcement wednesday as the european union announced \\
                it would soon undertake a partial privatization. \\
        \midrule
        \bottomrule
        {\bf (Title) france launches partial edf privatization} \\
    \end{tabular}
    \caption{Full text sequences generated by $f_{S\rightarrow F}^{(\text{Pr-Thr}), 1}$, $f_{S\rightarrow F}^{(\text{DBAE}), 1}$, and $f_{S\rightarrow F}^{(\bm{\mu}:1), 1}$ during the first back-translation loop.}
    \label{tab:art-examples}
\end{table*}



Let us consider the task of transforming a sequence in domain $A$ into a corresponding sequence in domain $B$ (e.g. sentences in two languages for machine translation). Let $\mathcal{D}_A$ and $\mathcal{D}_B$ be corpora of sequences in $A$ and $B$, without any mapping between their respective elements. The back-translation approach starts with initial seq2seq models $f^0_{A \rightarrow B}$ and $f^0_{B \rightarrow A}$, which can be hand-crafted or learned without aligned pairs, and uses them to create artificial aligned training data:
\begin{align}
    \mathcal{D}^0_{A' \rightarrow B} &= \Big \{ \big( f^0_{B \rightarrow A}(b), b \big); \;\; \forall b \in \mathcal{D}_B \Big \} \\
    \mathcal{D}^0_{B' \rightarrow A} &= \Big \{ \big( f^0_{A \rightarrow B}(a), a \big); \;\; \forall a \in \mathcal{D}_A \Big \} 
\end{align}
Let $\mathcal{S}$ denote a supervised learning algorithm, which takes a set of aligned sequence pairs and returns a mapping function. This artificial data can then be used to train the next iteration of seq2seq models, which in turn are used to create new artificial training sets ($A$ and $B$ can be switched here):
\begin{align}
    f^{i+1}_{A \rightarrow B} &= \mathcal{S}(\mathcal{D}^{i}_{A' \rightarrow B}) \\
    \mathcal{D}^{i+1}_{B' \rightarrow A} &= \Big \{ \big( f^{i+1}_{A \rightarrow B}(a), a \big); \;\; \forall a \in \mathcal{D}_A \Big \} 
\end{align}
The model is trained at each iteration on artificial inputs and real outputs, then used to create new training inputs. Thus, if the initial system isn't too far off, we can hope that training pairs get closer to the true data distribution with each step, allowing in turn to train better models.

In the case of summarization, we consider the domains of full text sequences $\mathcal{D}^F$ and of summaries $\mathcal{D}^S$, and attempt to learn \emph{summarization} ($f_{F\rightarrow S}$) and \emph{expansion} ($f_{S\rightarrow F}$) functions. However, contrary to the translation case, $\mathcal{D}^F$ and $\mathcal{D}^S$ are not interchangeable. Considering that a summary typically has less information than the corresponding full text, we choose to only define initial ${F\rightarrow S}$ models. We can still follow the proposed procedure by alternating directions at each step.

\subsection{Initialization Models for Summarization}

To initiate their process for the case of machine translation, \citet{DBLP:conf/emnlp/LampleOCDR18} use two different initialization models for their neural (NMT) and phrase-based (PBSMT) systems. The former relies on denoising auto-encoders in both languages with a shared latent space, while the latter uses the PBSMT system of \citet{DBLP:conf/naacl/KoehnOM03} with a phrase table obtained through unsupervised vocabulary alignment as in \citep{DBLP:journals/corr/abs-1805-11222}. 

While both of these methods work well for machine translation, they rely on the input and output having similar lengths and information content. In particular, the statistical machine translation algorithm tries to align most input tokens to an output word. In the case of text summarization, however, there is an inherent asymmetry between the full text and the summaries, since the latter express only a subset of the former. Next, we propose three initialization systems which implicitly model this information loss. Full implementation details are provided in the Appendix.

\paragraph{Procrustes Thresholded Alignment (Pr-Thr)} The first initialization is similar to the one for PBSMT in that it relies on unsupervised vocabulary alignment. Specifically, we train two skipgram word embedding models using \textsc{fasttext} \citep{DBLP:journals/tacl/BojanowskiGJM17} on $\mathcal{D}^F$ and $\mathcal{D}^S$, then align them in a common space using the Wasserstein Procrustes method of \citet{DBLP:journals/corr/abs-1805-11222}. Then, we map each word of a full text sequence to its nearest neighbor in the aligned space if their distance is smaller than some threshold, or skip it otherwise. We also limit the output length, keeping only the first $N$ tokens.
We refer to this function as $f_{F\rightarrow S}^{(\text{Pr-Thr}), 0}$.

\paragraph{Denoising Bag-of-Word Auto-Encoder (DBAE)} Similarly to both \citep{DBLP:conf/emnlp/LampleOCDR18} and \citep{DBLP:conf/emnlp/WangL18a}, we also devise a starting model based on a DAE. One major difference is that we use a simple Bag-of-Words (BoW) encoder with fixed pre-trained word embeddings, and a 2-layer GRU decoder. Indeed, we find that a BoW auto-encoder trained on the summaries reaches a reconstruction \textsc{rouge-l} f-score of nearly 70\% on the test set, indicating that word presence information is mostly sufficient to model the summaries. As for the noise model, for each token in the input, we remove it with probability $p/2$ and add a word drawn uniformly from the summary vocabulary with probability $p$.

The BoW encoder has two advantages. First, it lacks the other models' bias to keep the word order of the full text in the summary. Secondly, when using the DBAE to predict summaries from the full text, we can weight the input word embeddings by their corpus-level probability of appearing in a summary, forcing the model to pay less attention to words that only appear in $\mathcal{D}^F$. The Denoising Bag-of-Words Auto-Encoder with input re-weighting is referred to as $f_{F\rightarrow S}^{(\text{DBAE}), 0}$.

\paragraph{First-Order Word Moments Matching ($\bm{\mu}$:1)} We also propose an extractive initialization model. Given the same BoW representation as for the DBAE, function $f_\theta^\mu(s, v)$ predicts the probability that each word $v$ in a full text sequence $s$ is present in the summary. We learn the parameters of $f_\theta^\mu$ by marginalizing the output probability of each word over all full text sequences, and matching these first-order moments to the marginal probability of each word's presence in a summary. That is, let $\mathcal{V}^S$ denote the vocabulary of $\mathcal{D}^S$, then $\forall v \in \mathcal{V}^S$:
\begin{align*}
\mu_v^F = \frac{\sum_{s \in \mathcal{D}^F} \mathbbm{1}_{v \in s} }{|\mathcal{D}^F|} \;\; \text{and} \;\;  \mu_v^S = \frac{\sum_{s \in \mathcal{D}^s} \mathbbm{1}_{v \in s} }{|\mathcal{D}^S|}
\end{align*}
We minimize the binary cross-entropy (BCE) between the output and summary moments:
\begin{align*}
    \theta^* = \argmin \sum_{v \in \mathcal{V}^S} \text{BCE} \Big ( \frac{\sum_{s \in \mathcal{D}^F} f_\theta^\mu(s, v) }{|\mathcal{D}^F|}, \mu_v^S \Big )
\end{align*}
We then define an initial extractive summarization model by applying $f_{\theta^*}^\mu(\cdot, \cdot)$ to all words of an input sentence, and keeping the ones whose output probability is greater than some threshold. We refer to this model as $f_{F\rightarrow S}^{(\bm{\mu}:1), 0}$.

\subsection{Artificial Training Data}
\label{sec:art-method}

We apply the back-translation procedure outlined above in parallel for all three initialization models. For example, $f_{F\rightarrow S}^{(\bm{\mu}:1), 0}$ yields the following sequence of models and artificial aligned datasets:
\begin{align*}
& f_{F\rightarrow S}^{(\bm{\mu}:1), 0} \; \rightarrow \; \mathcal{D}_{S' \rightarrow F}^{(\bm{\mu}:1), 0} \; \rightarrow \; f_{S\rightarrow F}^{(\bm{\mu}:1), 1}  \; \rightarrow \; \mathcal{D}_{F' \rightarrow S}^{(\bm{\mu}:1), 1} \\
& \; \rightarrow \;  f_{F\rightarrow S}^{(\bm{\mu}:1), 2} \; \rightarrow \; \mathcal{D}_{S' \rightarrow F}^{(\bm{\mu}:1), 2} \; \rightarrow \; f_{S\rightarrow F}^{(\bm{\mu}:1), 3}  \; \rightarrow \; \ldots
\end{align*}
Finally, in order to take advantage of the various strengths of each of the initialization models, we also concatenate the artificial training dataset at each odd iteration to train a summarizer, e.g.:
\begin{align*}
f_{F\rightarrow S}^{(\text{All}), 2} &= \mathcal{S} \Big( \mathcal{D}_{F' \rightarrow S}^{(\text{Pr-Thr}), 1} \cup \mathcal{D}_{F' \rightarrow S}^{(\text{DBAE}), 1} \cup \mathcal{D}_{F' \rightarrow S}^{(\bm{\mu}:1), 1} \Big)
\end{align*}

\begin{table}[t]
    \centering
    \begin{tabular}{l|c|c|c}
                & \textsc{R-1}  & \textsc{R-2}  & \textsc{R-L} \\
    \toprule
    \bottomrule
    Lead-8      & 21.86         & 7.66          & 20.45 \\
    PBSMT       & 24.29         & 8.65          & 21.82 \\      
    Pre-DAE$^1$ & 21.26         & 5.60          & 18.89 \\
    \midrule
    (Pr-Thr)-0    & 24.79         & \textbf{8.80} & 22.46 \\
    (DBAE)-0      & 28.58         & 6.74          & 22.72 \\
    ($\bm{\mu}$:1)-0 & \textbf{29.17}& 8.10          & \textbf{24.71} \\
    \end{tabular}
    \caption{Test \textsc{rouge} for trivial baseline and initialization systems. $^1$\citep{DBLP:conf/emnlp/WangL18a}.}
    \label{tab:base-bootstrap}
\end{table}

\section{Experiments}
\label{sec:experiments}

\paragraph{Data and Model Choices} We validate our approach on the Gigaword corpus, which comprises of a training set of 3.8M article headlines (considered to be the full text) and titles (summaries), along with 200K validation pairs, and we report test performance on the same 2K set used in \citep{DBLP:conf/emnlp/RushCW15}. Since we want to learn systems from fully unaligned data without giving the model an opportunity to learn an implicit mapping, we also further split the training set into  2M examples for which we only use titles, and 1.8M for headlines.
All models after the initialization step are implemented as convolutional seq2seq architectures using Fairseq \citep{DBLP:journals/corr/abs-1904-01038}. Artificial data generation uses top-15 sampling, with a minimum length of 16 for full text and a maximum length of 12 for summaries. \textsc{rouge} scores are obtained with an output vocabulary of size 15K and a beam search of size 5 to match  \citep{DBLP:conf/emnlp/WangL18a}.

\paragraph{Initializers} Table~\ref{tab:base-bootstrap} compares test \textsc{ROUGE} for different initialization models, as well as the trivial Lead-8 baseline which simply copies the first 8 words of the article. We find that simply thresholding on distance during the word alignment step of (Pr-Thr) does slightly better then the full PBSMT system used by \citet{DBLP:conf/emnlp/LampleOCDR18}. Our BoW denoising auto-encoder with word re-weighting also performs significantly better than the full seq2seq DAE initialization used by \citet{DBLP:conf/emnlp/WangL18a} (Pre-DAE). The moments-based initial model ($\bm{\mu}$:1) scores higher than either of these, with scores already close to the full unsupervised system of \citet{DBLP:conf/emnlp/WangL18a}.

In order to investigate the effect of these three different strategies beyond their  \textsc{rouge} statistics, we show generations of the three corresponding first iteration expanders for a given summary in Table~\ref{tab:art-examples}. The unsupervised vocabulary alignment in (Pr-Thr) handles vocabulary shift, especially changes in verb tenses (summaries tend to be in the present tense), but maintains the word order and adds very little information. Conversely, the ($\bm{\mu}$:1) expansion function, which is learned from purely extractive summaries, re-uses most words in the summary  without any change and adds some new information. Finally, the auto-encoder based (DBAE) significantly increases the sequence length and variety, but also strays from the original meaning (more examples in the Appendix). The decoders also seem to learn facts about the world during their training on article text (EDF/GDF is France's public power company). 

\paragraph{Full Models} Finally, Table~\ref{tab:full-res} compares the summarizers learned at various back-translation iterations to other unsupervised and semi-supervised approaches. Overall, our system outperforms the unsupervised Adversarial-\textsc{reinforce} of \citet{DBLP:conf/emnlp/WangL18a} after one back-translation loop, and most semi-supervised systems after the second one, including \citet{DBLP:journals/corr/SongTQLL19}'s MASS pre-trained sentence encoder and \citet{DBLP:conf/emnlp/MiaoB16}'s Forced-attention Sentence Compression (FSC), which use 100K and 500K aligned pairs respectively.
As far as back-translation approaches are concerned, we note that the model performances are correlated with the initializers' scores reported in Table~\ref{tab:base-bootstrap} (iterations 4 and 6 follow the same pattern). In addition, we find that combining data from all three initializers before training a summarizer system at each iteration as described in Section~\ref{sec:art-method} performs best, suggesting that the greater variety of artificial full text does help the model learn.

\begin{table}[t!]
    \centering
    \begin{tabular}{l|c|c|c|c}
                        & Sup.  & \textsc{R-1}  & \textsc{R-2}  & \textsc{R-L} \\
    \toprule
    \bottomrule
    (Pr-Thr)-2          & 0     & 26.17         & 9.42          & 23.65 \\
    (DBAE)-2            & 0     & 28.55         & 10.24         & 25.46 \\
    ($\bm{\mu}$:1)-2    & 0     & 29.55         & 9.62          & 26.10 \\
    (All)-2             & 0     & 29.80         & 11.52         & 27.01 \\
    (All)-4             & 0     & {\bf 30.19}   & 12.36   & {\bf 27.75} \\
    (All)-6             & 0     & 30.04         & {\bf 12.69}   & 27.64 \\
    \midrule
    Advers.             & 0     & 28.11 & 9.97          & 25.41 \\
    \textsc{rein-}      & 10K   & 30.01         & 11.57         & 27.61 \\
    \textsc{force}$^1$  & 500K  & 33.33         & 14.18         & 30.48 \\
    \midrule
    MASS$^2$            & 100K  & 29.79         & 12.75         & 27.45 \\
    \midrule
    FSC$^3$             & 500K  & 30.14         & 12.05         & 27.99 \\
    \bottomrule
    \midrule
    Seq2seq$^4$         & 3.8M  & 35.30         & 16.64         & 32.62 \\
    \end{tabular}
    \caption{Comparison of full systems. The best scores for unsupervised training are bolded. Results from: $^1$\citep{DBLP:conf/emnlp/WangL18a}, $^2$\citep{DBLP:journals/corr/SongTQLL19}, $^3$\citep{DBLP:conf/emnlp/MiaoB16}, and $^4$\citep{DBLP:conf/conll/NallapatiZSGX16}} 
    \label{tab:full-res}
\end{table}

\paragraph{Conclusion} In this work, we use the back-translation paradigm for unsupervised training of a summarization system. We find that the model benefits from combining initializers, matching the performance of semi-supervised approaches. 

\clearpage

\bibliography{unsup_sum_emnlp}
\bibliographystyle{acl_natbib}

\clearpage

\appendix

\section{Implementation Choices for Initialization and Seq2seq Models}

We describe the modeling choices for initialization models (Pr-Thr), (DBAE), and ($\bm{\mu}$:1). All hyper-parameters for each of these systems are set based on the models' \textsc{rouge-l} score on the validation set. Unless otherwise stated, all models use Skipgram FastText\footnote{\url{https://fasttext.cc/}} word embeddings which are shared across the input and output layers. The dimension 512 embeddings are trained on the concatenation of the full text and summary sequences ${\mathcal{D}^F \cup \mathcal{D}^S}$. $\mathcal{V}$ is the full vocabulary, and $\mathcal{V}^F$ and $\mathcal{V}^S$ are the vocabularies of $\mathcal{D}^F$ and $\mathcal{D}^S$ respectively. All trained models use the Adam optimizer with learning rate $5e-4$. The convolutional seq2seq models use the \emph{fconv\_iwslt\_de\_en} architecture previded in Fairseq\footnote{\url{https://fairseq.readthedocs.io/en/latest/models.html}} with pre-trained input and output word embeddings, a vocabulary size of 50K for the full text and of 15K for the summaries. For the expander generations, we collapse contiguous UNK tokens, and cut the sentence at the first full stop even when the model did not generate an EOS token, yielding outputs that are sometimes shorter than 16 words.

\paragraph{Procrustes Thresholded Alignment (Pr-Thr)} For this model, we train two sets of word embeddings on $\mathcal{D}^F$ and $\mathcal{D}^S$ separately, and compute aligned vectors using the FastText implementation of the \citep{DBLP:journals/corr/abs-1805-11222} algorithm\footnote{\url{https://github.com/facebookresearch/fastText/tree/master/alignment}}. We then map each word in an input sequence to its closest word in $\mathcal{V}^S$ in the aligned space, unless the nearest neighbor is the EOS token or the distance to the nearest neighbor in the aligned space is greater than a threshold $\eta$. The output sequence then consists in the first $N$ mapped words in the order of the input sequence. We found that using embeddings of dimension $256$, threshold $\eta = 0.9$, and maximum output length $N = 12$ yields the best validation \textsc{rouge-l}.

We compare (Pr-Thr) to a PBSMT baseline in Table~\ref{tab:base-bootstrap}. We use the UnsupervisedMT codebase\footnote{\url{https://github.com/facebookresearch/UnsupervisedMT/tree/master/PBSMT}} of \citep{DBLP:conf/emnlp/LampleOCDR18} with the same pre-trained embedding, and also perform a hyper-parameter search over maximum length, which sets $N=15$.

\paragraph{Denoising Bag-of-Word Auto-Encoder (DBAE)} The DBAE is trained on all sentences in $\mathcal{D}^S$. The encoder of the DBAE averages the input word embeddings and applies a linear transformation, followed by a Batch Normalization layer \citep{DBLP:conf/icml/IoffeS15}. The decoder is a 2-layer GRU recurrent neural network with hidden dimension 256. The encoder output is concatenated to the initial hidden state of both layers, then projected back down to the hidden dimension. 

To use the model for summarization, we perform two changes from the auto-encoding setting. First, we perform a weighted instead of a standard average, where words that are less likely to appear in $\mathcal{D}^S$ than in $\mathcal{D}^F$ are down-weighted (and words that are in $\mathcal{V}^F$ but not in $\mathcal{V}^S$ are dropped). Specifically, given a word $v \in \mathcal{V}^S$, its weight $w_v$ in the summarization weighted BoW encoder is given as:
\begin{align}
\mu_v^F = \frac{\sum_{s \in \mathcal{D}^F} \mathbbm{1}_{v \in s} }{|\mathcal{D}^F|} \;\; \text{and} \;\;  \mu_v^S = \frac{\sum_{s \in \mathcal{D}^s} \mathbbm{1}_{v \in s} }{|\mathcal{D}^S|} \label{eq:moments_apdx}
\end{align}
\begin{align}
    w_v = \max(\frac{\mu_v^S}{\mu_v^F}, 1)
\end{align}
Secondly, we implement something like a pointer mechanism by adding $\lambda$ to the score of each of the input words in the output of the GRU, before the softmax. At test time and when creating artificial data, we decode with beam search and a beam size of size 5, maximum output length $N=15$, and input word bias $\lambda=2$.

\paragraph{First-Order Word Moments Matching ($\bm{\mu}$:1)} The moments matching model uses the same encoder as the (DBAE) followed by a linear mapping to the summary vocabulary, followed by a sigmoid layer (the log-score of all words that do not appear in the input is set to $-1e6$). Unfortunately, computing the output probabilities for all sentences in the corpus before computing the Binary Cross-Entropy is impractical, and so we implement a batched version of the algorithm. Let corpus-level moments $\mu_v^F$ and $\mu_v^S$ be defined as in Equation~\ref{eq:moments_apdx}. Let $\mathcal{B}^F$ be a batch of full text sequences, we define:
\begin{align}
\hat{\mu}_v^F = \frac{\sum_{s \in \mathcal{B}^F} \mathbbm{1}_{v \in s} }{|\mathcal{B}^F|} \;\; \text{and} \;\;  \hat{\mu}_v^S = \frac{\hat{\mu}_v^F }{\mu_v^F } . \mu_v^S
\end{align}
For each batch, the algorithm then takes a gradient step for the loss:
\begin{align*}
    \hat{\mathcal{L}}(\mathcal{B}^F; \theta) = \sum_{v \in \mathcal{V}^S} \text{BCE} \Big ( \frac{\sum_{s \in \mathcal{B}^F} f_\theta^\mu(s, v) }{|\mathcal{D}^F|}, \hat{\mu}_v^S \Big )
\end{align*}
The prediction is similar as for the (Pr-Thr) system except that we threshold on $f_\theta^\mu(s, v)$ rather than on the nearest neighbor distance, with threshold ${\eta=0.3}$ (the maximum output length is also ${N=12}$)

\begin{table*}
    \centering
    \begin{tabular}{l}
        {\bf over N,NNN ancient graves found in greek metro dig} \\
        \toprule
        \bottomrule
        (Pr-Thr)  over N,NNN ancient graves were found in a greek metro -lrb- UNK -rrb-. \\
        \midrule 
        (DBAE)    the remains of N,NNN graves on ancient greek island have been found in three ancient  \\
                graves in the past few days, a senior police officer said on friday. \\
        \midrule
        ($\bm{\mu}:1$) over N,NNN ancient graves have been found in the greek city of alexandria in the northern \\
                greek city of salonika in connection with the greek metro and dig deep underground.  \\
    \multicolumn{1}{c}{} \\
        {\bf  ukraine's crimea dreams of union with russia} \\
        \toprule
        \bottomrule
        (Pr-Thr)  ukraine 's crimea UNK of the union with russia. \\
        \midrule 
        (DBAE)    ukraine has signed two agreements with ukraine on forming its european union and \\
                ukraine as its membership.  \\
        \midrule
        ($\bm{\mu}:1$) ukraine's crimea peninsula dreams of UNK, one of the soviet republic's most UNK country \\
                with russia, the itar-tass news agency reported. \\
    \multicolumn{1}{c}{} \\
        {\bf  malaysian opposition seeks international help to release detainees} \\
        \toprule
        \bottomrule
        (Pr-Thr) the malaysian opposition thursday sought international help to release detainees. \\
                the malaysian opposition, news reports said.  \\
        \midrule 
        (DBAE)   malaysian prime minister abdullah ahmad badawi said tuesday that the government's  \\
                decision to release NNN detainees, a report said wednesday. \\
        \midrule
        ($\bm{\mu}:1$) malaysian opposition parties said tuesday it seeks to ``help'' the release of detainees. \\
    \multicolumn{1}{c}{} \\
        {\bf  russia to unify energy transport networks with georgia rebels} \\
        \toprule
        \bottomrule
        (Pr-Thr)  russia is to unify energy transport networks with georgia rebels. \\
        \midrule 
        (DBAE)    russian government leaders met with representatives of the international energy giant said \\
                monday that their networks have been trying to unify their areas with energy supplies. \\
        \midrule
        ($\bm{\mu}:1$) russia is to unify its energy and telecommunication networks to cope with georgia's \\
                    separatist rebels and the government.  \\
    \multicolumn{1}{c}{} \\
        {\bf  eu losing hope of swift solution to treaty crisis} \\
        \toprule
        \bottomrule
        (Pr-Thr)  the eu has been losing hope of a UNK solution to the maastricht treaty crisis. \\
        \midrule 
        (DBAE)    the european union is losing hope it will be a swift solution to the crisis of the eu \\
                -lrb- eu -rrb-, hoping that it's in an ``urgent'' referendum. \\
        \midrule
        ($\bm{\mu}:1$) eu governments have already come under hope of a swift solution to a european union treaty \\
                that ended the current financial crisis. \\
    \end{tabular}
    \caption{More examples of artificial data after the first back-translation iteration.}
    \label{tab:examples_more}
\end{table*}

\begin{table*}[t]
    \centering
    \begin{tabular}{l}
         (Original) \emph{malaysia has drafted its first legislation aimed at punishing computer hackers,} \\ 
         \emph{an official said wednesday.} \\
         \midrule
         \bottomrule
        (Pr-Thr)-1 malaysia has enacted a draft, the first law on a UNK computer hacking.  \\
        (Pr-Thr)-3 malaysia has issued a draft of the law on computer hacking. \\
        (Pr-Thr)-5 malaysia has drafted a first law on the computer hacking and internet hacking.  \\
        \midrule
        \bottomrule
        (DBAE)-1   malaysia's parliament friday signed a bill to allow computer users to  \\
        monitor UNK law.  \\
        (DBAE)-3  the country has been submitted to parliament in NNNN passed a bill wednesday \\
        in the first reading of the computer system, officials said monday. \\
        (DBAE)-5  malaysia's national defense ministry has drafted a regulation of computer  \\
        hacking in the country, the prime minister said friday. \\
        \midrule
        ($\bm{\mu}:1$)-1 malaysia will have drafts the first law on computer hacking. \\
        ($\bm{\mu}:1$)-3 malaysia has started drafts to be the first law on computer hacking. \\
        ($\bm{\mu}:1$)-5 malaysia today presented the nation's first law on computer hacking in the \\
         country, news reports said wednesday. \\
        \midrule
        \bottomrule
        {\bf (Title) malaysia drafts first law on computer hacking} \\
    \end{tabular}
    \caption{Evolution of generated full text sequences across iterations.}
    \label{tab:examples_evolve}
\end{table*}

\begin{table*}[t]
    \centering
    \begin{tabular}{l}
        (Article) \emph{chinese permanent representative to the united nations wang guangya on wednesday urged} \\
         \emph{the un and the international community to continue supporting timor-leste.} \\
        (Pred) chinese permanent representative urges un to continue supporting timor-leste \\
        (Title) china stresses continued international support for timor-leste \\
        \midrule
        (Article) \emph{macedonian president branko crvenkovski will spend orthodox christmas this weekend with } \\
         \emph{the country's troops serving in iraq, his cabinet said thursday.} \\
        (Pred) macedonian president to spend orthodox christmas with troops in iraq \\
        (Title) macedonian president to visit troops in iraq \\
        \midrule
        (Article) \emph{televangelist pat robertson, it seems, isn't the only one who thinks he can see god's } \\
         \emph{purpose in natural disasters.} \\
        (Pred) evangelist pat robertson thinks he can see god's purpose in disasters \\
        (Title) editorial: blaming god for disasters \\
        \midrule
        (Article) \emph{the sudanese opposition said here thursday it had killed more than NNN government } \\
         \emph{soldiers in an ambush in the east of the country.} \\
        (Pred) sudanese opposition kills N government soldiers in ambush \\
        (Title) sudanese opposition says NNN government troops killed in ambush \\
    \end{tabular}
    \caption{Example of model predicitons for $f_{F \rightarrow S}^{(\text{All}, 6)}$.}
    \label{tab:examples_prediction}
\end{table*}

\section{More Examples of Model Predictions}

We present more examples of the expander and summarizer models' outputs in Tables~\ref{tab:examples_more}, \ref{tab:examples_evolve}, and \ref{tab:examples_prediction}. Table~\ref{tab:examples_more} shows more expander generations for all three initial models after one back-translation epoch. They follow the patterns outlined in Section~\ref{sec:experiments}, with (DBAE) showing more variety but being less faithful to the input. Table~\ref{tab:examples_evolve} show generations from the expander models at different back-translation iteration. It is interesting to see that each of the three models slowly overcome their initial limitations: the (DBAE) expander's third version is much more faithful to the input than its first, while the moments-based approach starts using rephrases and modeling vocabulary shift. The Procrustes method seems to benefit less from the successive iterations, but still starts to produce longer outputs. Finally, Table~\ref{tab:examples_prediction} provides summaries produced by the final model. While the model does produce likely summaries, we note that aside from the occasional synonym use or verbal tense change, and even though we do not use an explicit pointer mechanism beyond the standard seq2seq attention, the model's outputs are mostly extractive.

\end{document}